\documentclass[a4paper, 11pt]{article}
\pdfoutput=1

\usepackage{amssymb,amsmath}
\usepackage{graphicx}

\usepackage{color,xcolor}
\usepackage{wrapfig,rotating}
\usepackage{subfigure}
\usepackage{array}
\usepackage{xspace}
\usepackage{dsfont}
\usepackage{rotating}
\usepackage[noend]{algorithmic}
\usepackage{multirow}
\usepackage[vlined,linesnumbered,ruled,titlenotnumbered]{algorithm2e}

\newcommand{\ignore}[1]{}

\begin{document}

\title{A Fast and Effective Local Search Algorithm for Optimizing the Placement of Wind Turbines}

\author{Markus Wagner, Jareth Day, and Frank Neumann\\
School of Computer Science, University of Adelaide,\\Adelaide, SA 5005, Australia}
\date{}
\maketitle

\begin{abstract}
The placement of wind turbines on a given area of land such that the wind farm produces a maximum amount of energy is a challenging optimization problem. In this article, we tackle this problem, taking into account wake effects that are produced by the different turbines on the wind farm.
We significantly improve upon existing results for the minimization of wake effects by developing a new problem-specific local search algorithm. One key step in the speed-up of our algorithm is the reduction in computation time needed to assess a given wind farm layout compared to previous approaches. Our new method allows the optimization of large real-world scenarios within a single night on a standard computer, whereas weeks on specialized computing servers were required for previous approaches.

\end{abstract}


\section{Introduction}

Renewable energy plays an increasing role in the energy supply world-wide, and wind energy is one of the key players in the field of renewable energy. 
The \emph{wind farm layout problem} entails the process of planning the placement of turbines (and supporting equipment) on a potential wind farm site. 
The creation of a farm layout involves the invocation of a software optimization module, which attempts to efficiently place the turbines while adhering to the constraints and optimizing the stated objectives. Often, this module is embedded within a specialized tool provided by wind power consultants such as Garrad Hassan or AWS TruePower, who offer a product such as OpenWind~\cite{openwind}. 

One of the problems that such tools have to deal with, apart from the actual optimization, is the scaling cost of computing wake effects when estimating energy capture for increasing numbers of turbines. To estimate the energy capture of a layout the optimization module models the free stream wind flowing through the site in and out of the turbines. Some degree of non-linear wind turbulence occurs at the outflow of a turbine and affects the inflow to turbines close enough behind it. Modelling this effect is necessary because wake has a great effect on the actual energy output of a wind farm. However, the time for computing the wake effect with respect to a given wake model such as the Park wake model~\cite{modParkWakeModel} takes time $\Omega(n^2)$, where $n$ is the number of turbines on the wind farm. This computational effort is significant if one is applying iterative search algorithms such as local search, simulated annealing or evolutionary algorithms for the optimization of the placement. Such methods would need to evaluate each farm layout with respect to the wake model under consideration and would therefore require time $\Omega(n^2)$ for each solution that is considered during the optimization process.

\subsection{Related Work}\label{sect:relatedwork}

The optimal siting of wind turbines on a given area of land is a complex optimization problem which is hard to solve by exact methods.
The decision space is non-linear with respect to how sited turbines interact, when considering wake loss and energy capture. 
Several bio-inspired computation techniques such as evolutionary algorithms~\cite{EibSmi2007,goldberg:1989:book} and particle swarm optimization~\cite{kennedyEberhart:book} have been used for the optimization.  

The different approaches for the wind farm layout problem are summarized in~\cite{stateofart2010}.
Wan et al, \cite{wanEtAlBinaryGA,wanEtAlRCGA,wanEtAlPSO}, use cell-based approaches and compare different bio-inspired algorithms, each applied to the same set of wind farm models and parameters. They use successively more expressive layout representations 
to relax where in a cell a turbine can be located: strictly in the middle, anywhere, or anywhere subject to proximity constraints with neighbouring turbines.
An alternative to cell placement was explored in \cite{KusiakLayout2010}: each turbine's location is a decision variable pair of real-valued, spatial (x,y) coordinates. In that paper a simple evolution strategy (ES) is applied to optimize the placement of the turbines. In general, an ES is effective because it is easily parallelized and it \emph{self-adapts} the extent to which it perturbs decision variables when generating a new potential solution. 
However, the algorithm given in \cite{KusiakLayout2010} is only able to deal with problems that have just a small number of turbines.
A more powerful ES called CMA-ES has been used for the same problem in~\cite{EWEA2011}. 
The advantage of this approach is that it allows for the effective optimization of huge layouts for up to 1000 turbines. 
Nevertheless, this approach is still computationally expensive, requiring up to weeks to solve for large wind farm layouts on computing servers.

Many of the above-mentioned approaches make simplified assumptions regarding the realism of the wake models and the wind scenarios. Even in recent work, it is either completely ignored, assumed to be constant, or highly simplified wind scenarios are used (except for~\cite{KusiakLayout2010,EWEA2011}). 
For example, in the optimization formulation in~\cite{mustakerov2011}, the turbines are placed in regular grids, and wake is considered by enforcing minimal and maximal distances between the turbines, thus effectively neglecting physical effects. 
In \cite{tzanos2011}, distance-dependent wakes are considered, but the underlying wind scenario was randomly generated, and is the result of unrealistic assumptions. 
On the other hand, the above-mentioned industry tool AWS OpenWind contains elaborate wake models that are used for real-world scenarios, but its simple optimizer neither considers the turbines' vicinities when perturbing layouts, nor adapts parameters for the perturbation during the run.

Note that, placing turbines on a defined area is loosely related to the difficult problem of packing discs in shapes~\cite{GrahamLNO98,Szabo2007}. There, the task is to arrange $n$ identical discs without overlap entirely inside a square on the plane, such that the discs have the largest possible diameter. As the turbines influence each other via non-linear wake-effects, these ``influence areas" can be assumed to be circular for very simple scenarios of uniform wind distributions.\footnote{Note that these discs are not necessarily of infinite size, as a cut-off distance can be set once the influence becomes negligible.} The naive goal then can be to minimize the overlap of the influence discs. However, the theoretical results from the disc packing are not applicable in our context, as it would be extremely simplifying to assume that these discs are circular due to non-uniform wind distributions. Additionally, the sizes and irregular shapes are highly dependent on the interaction of the turbines: due to interactions, the wind resources available at a turbine change, and so does then the influence area around a turbine.

\subsection{Our Contribution}
In this paper we describe the design of a fast and effective randomized local search algorithm to act as a farm layout optimization module. The algorithm

\begin{enumerate}
\item uses a fast approach for evaluating new layouts, 
\item makes effective use of the problem characteristics to produce new layouts, and
\item can efficiently handle infeasible regions in order to respect geographical constraints.
\end{enumerate}

To be precise, where previous algorithms required time $\Omega(n^2)$ for the computation of the wake effects within a new layout, our algorithm requires $\Theta(kn)$, where $n$ is the number of turbines on the wind farm, and $k$ the number of turbines whose location was changed over the previous layout.

The new local search algorithm is capable of generating good layouts for hundreds of turbines quickly on standard computers, without the need of specialised computing hardware. 
This allows it to accommodate the emerging requirements of larger farms. For example, the Horse Hollow Wind Energy Center in Texas, USA operates with 735.5 megawatt capacity and consists of more than 300 turbines spread over nearly 47,000 acres (190 km$^{2}$). 
We compare our new algorithm to two state-of-the-art approaches: the CMA-ES approach presented in~\cite{EWEA2011} and the local search approach used in the industry tool AWS OpenWind. 
Our experiments show that the new algorithm outperforms both approaches in terms of the quality of the results and the running time.

We proceed as follows: Section~\ref{sect:layoutOpt} provides a description of the farm layout optimization problem. For this, it is necessary to explain how wake affects wind characteristics as wind propagates through the site, and how the expected energy capture is calculated. 
Section~\ref{sect:rls} describes our algorithm. Section~\ref{sect:results} gives details about our experiments, and discusses the outcomes. 
Then, the capability to deal with infeasible areas is shown, using a scenario based on an existing near-shore wind farm.
Finally, Section~\ref{sect:conclusion} summarizes our findings.


\section{Wind Farm Layout Optimization}\label{sect:layoutOpt}

Let $X=\{x_1, \ldots, x_n\}$ and $Y=\{y_1, \ldots, y_n\}$ be a set of $x$ and $y$ coordinates of $n$ wind turbines in the plane. 
The goal is to find a set of coordinates such that the energy output of the whole wind farm is maximized. Furthermore, the layout has to respect certain constraints.

\setlength{\algomargin}{1em}
\begin{figure}[t]
\begin{minipage}[t]{\textwidth}
\begin{algorithm}[H]
\caption{Procedure for evaluation of wake effects due to the Park model \cite{KusiakLayout2010}}
\label{alg:ParkModel}
Given $\{X,Y\}$ as turbine locations, turbine thrust coefficient $C_T$, rotor diameter $R$, landscape-specific wake spreading factor $\kappa$ \; 
$a=1-\sqrt{1-C_T}$, $b=\kappa/R$, $u \Leftarrow $ unit step function, $o={(x_i-x_j)cos \theta + (y_i-y_j)sin \theta}$\;
$d_{i,j}=\|o\|$, $\alpha=tan^{-1}\kappa$\;
        \For{$ i=1$ to number of turbines}	{
\For{$\theta=0^0$ to $360^0$}{
\For{$j=1$ to n-1  and $j \neq i$}{
$\delta_{i,j}=cos^{-1}\left({\frac{o+R/\kappa}{\sqrt{(x_i-x_j+\frac{R}{\kappa} cos\theta)^2+(y_i-y_j+\frac{R}{\kappa} sin \theta)^2}}}\right)$\;
$Vdef_{(i,j)}=u(\delta_{i,j}-\alpha)\frac{a}{(1+bd_{i,j})^2}$\;
}
$Vdef_{i}^{\theta}=\sqrt{\sum_{j} (Vdef_{(i,j)}^{\theta})^2}$\;
$c_i(\theta)=c_i(\theta) \times (1-Vdef_{i})$\;
}}
\end{algorithm}
\end{minipage}
\end{figure}

\subsection{Energy Output}
Depending on the chosen coordinates the overall energy output of the wind farm varies as we have to take the wake effects into account. 

We consider the Park wake model. In this model the wake effects on a turbine $i$ change the wind resource available to it along different directions by reducing the \textit{scale} parameter $c$ of the Weibull distribution estimated for the entire farm, which is also called the freestream wind resource. This is dependent on its location and the location of the rest of the turbines.
Hence, we have a parameter $c_i$ for each turbine $i$: its computation is complex and involves wind velocity deficits $Vdef_j$ that the turbine $i$ experiences due to the influence of other turbines $j$, $j \neq i$ (see Algorithm~\ref{alg:ParkModel}). We refer the reader to \cite{KusiakLayout2010} for a detailed presentation on the computation of this parameter when considering wake effects in the Park wake model. 
The expected energy output $\eta$ of the whole wind farm is given by 
\begin{equation}\label{eq:totalenergy}
E^{farm}[\eta]=\sum_i \int_\theta P(\theta) \int_v p(v(\theta),c_i(\theta,X,Y),k(\theta)) \beta^i(v).
\end{equation}

In this equation $v$ is the wind speed, and the function $\beta^i(v)$ defines the power curve for turbine $i$. Wind speed $v$ however is a random variable with a Weibull distribution, $p(v(\theta),c_i(\theta,X,Y),k(\theta))$, which is estimated from wind resource data and considers the wake effect using $X$ and $Y$. This distribution is also a function of the wind direction, $\theta$ which varies from $0^{\small{0}} - 360^{\small{0}}$. Note that the \emph{shape parameter} of the Weibull distribution is not influenced by the Wake effects here. 
Additionally, wind flows from a certain direction with some probability $P(\theta)$. 

\subsection{Constraints and Assumptions}\label{sec:constraints}
We have the following constraints placed on our optimization function. 
The first one enforces an upper bound on the area of the farm. This constraint ensures that we can only place a turbine $i$ within a certain area, which is a realistic constraint for most layout problems. 
For a rectangular farm 
with length $l$ and width $w$ this constraint is satisfied \textit{iff}
$0 \leq x_i \leq l~\&~0 \leq y_i \leq w, \forall{i}.$ 

The second constraint regulates the spatial proximity, as it dictates the minimal distance within which two turbines can be set up. It is satisfied \textit{iff}
$\sqrt{(x_i-x_j)^2+(y_i-y_j)^2} \geq MR, \forall i \forall j$ 
where $R$ is the rotor radius and $M$ is a proximity factor usually decided ahead of the optimization based on the make and model of the turbines used. 
We use $M=8$ based on the industry standard.

In addition to the above constraints, we assume that all turbines have the same power curves (approximated as piecewise linear functions) and that the same wind resource spans the entire farm.\footnote{To increase accuracy, these resources can be estimated for different parts in the farm.} The assumptions can be revised in a very straight forward manner to generate more realistic scenarios.


\section{Turbine Distribution Algorithm}\label{sect:rls}

In this section, we describe our algorithm for the optimization of wind farm layouts. It includes a problem specific local search operator and a faster evaluation.

\subsection{Computational Speedup}

First, we describe how computation time can be saved, if the new layout differs from the old one only in the location of a single turbine.

For the computation of the (turbine-specific) wind resources, 
the Weibull scale parameters are adjusted up to $24 \times n^2$ times (line 10 of Algorithm~\ref{alg:ParkModel}), as the wind direction is discretized into 24 wind directions in our model, and the mutual influence of the $n$ turbines has to be considered. 
The computation of the influence matrix results in an evaluation time that is quadratic with the number of turbines, causing evaluation times of up to 30 seconds on standard hardware for a 1000 turbine layout.\footnote{This test and all subsequent experiments were performed on AMD Opteron 250 CPUs (2.4GHz), on Debian GNU/Linux 5.0.8, with Java SE RE 1.6.}

As our optimization algorithm modifies only a single turbine of the current layout, the evaluation can be speeded up by updating the velocity deficits in line 10 of  Algorithm~\ref{alg:ParkModel} intelligently. 
First, for a moved turbine $i$ the influence of all other turbines on $i$ has to be computed conventionally, requiring $n$-1 influence checks. 
Second, the other turbines' velocity deficits have to be updated as $i$'s influence on these may have changed. This can be done as follows. For each such unmoved turbine $j$:
\begin{eqnarray}
Vdef_{j}^{\theta, \mathit{NEW}} = 
\sqrt{
\left( Vdef_{j}^{\theta, \mathit{OLD}} \right)^2 - 
\left( Vdef_{(j,i)}^{\theta, \mathit{OLD}} \right)^2 +
\left( Vdef_{(j,i)}^{\theta, \mathit{NEW}} \right)^2
\label{equ:fastUpdate}
}
\end{eqnarray}
\normalsize

\noindent where $Vdef_{j}^{\theta, \mathit{OLD}}$ is the velocity deficit the turbine $j$ experiences in the old layout, and $Vdef_{(j,i)}^{\theta, \mathit{OLD}}$ is the influence that $i$ had on $j$, for a given wind direction~$\theta$.

Over the course of the algorithm's run, we only have to initially create the three dimensional array carrying all mutual influences for all wind directions once. In subsequent evaluations, we can use and update this matrix to speedup the evaluations. Generalized, this allows for a resulting runtime for such a subsequent evaluation of only $\Theta(kn)$, when the locations of $k$ out of $n$ turbines were changed.\footnote{We use $k=1$ in our experiments, which results in a significant speedup (see Table~\ref{tab:results}).}

\subsection{The Algorithm}\label{sec:tda}
It is clear that the constraints discussed in Section \ref{sec:constraints} are vital to the construction of the algorithm. To ensure constraint handling in the optimization, a random local search algorithm was purpose-built for the application. The turbine distribution algorithm (TDA) described in Algorithm~\ref{alg:TurbineAlgorithm} iteratively displaces a single turbine in order to increase the energy gain while ensuring constraints are upheld.

\setlength{\algomargin}{1.5em}
\begin{figure}[p]
\begin{minipage}[t]{\textwidth}
\begin{algorithm}[H]
Given number of turbines $n$, map bounds $x_{max}$, $y_{max}$, nearest neighbours $nn$, reversal probability $p$, displacement distance standard deviation ${\sigma_{\{dis\}}}_{k}$ for ${k}\in\{1,\ldots, n\}$, and direction standard deviation ${\sigma_{\{dir\}}}_{k}$ for ${k}\in\{1,\ldots, n\}$, place $\{X,Y\} = \{[0, \ldots, x_{max}], [0, \ldots, y_{max}] \}^n$ in the grid formation of greatest space\;
Determine $f(\{X,Y\})$ as per Equation \ref{eq:totalenergy}\;
\For{$ i=1$ to number of evaluations}{
Set $\{X,Y\}' = \{X,Y\}$\;
Select turbine $k$ uniformly at random for the $i^{th}$ modification, denote as $\{x'_k,y'_k\}$\;
Determine nearest neighbours: $\{x''_1,y''_1\}, ..., \{x''_{nn}, y''_{nn}\}$\;
$\vec v = \left(\sum^{nn}_{j=1} x'_k - x''_j, \sum^{nn}_{j=1} y'_k - y''_j \right)$\;
$\vec v = \vec v/||\vec v||$\;
Select $\theta$ normally distributed over $\mu=\angle\vec v$ and $\sigma={\sigma_{\{\mathit{dir}\}}}_{k}$\;
Select $d$ normally distributed over $\mu=0$ and $\sigma = {\sigma_{\{\mathit{dis}\}}}_{k}$\;
$\vec v' = \theta * d$\;
Set $\vec v'$=$-\vec v'$ with probability $p$, as tighter groups may increase the farm's output\;
If applying $\vec v'$ would place $\{x'_k,y'_k\}$ in illegal area\hspace{-.1mm}, reduce length of $\vec v'$ until legal\;
Displace $\{x'_k,y'_k\}$ by $\vec v'$\;
Determine $f(\{X,Y\}')$ as per Equations \ref{eq:totalenergy} and \ref{equ:fastUpdate}\;
\If{$f(\{X,Y\}) \le f(\{X,Y\}')$}{
Set $\{X,Y\} = \{X,Y\}'$\;
Increase ${\sigma_{\{dis\}}}_{k}$\;}
\Else{Decrease ${\sigma_{\{dis\}}}_{k}$\;}
}
\caption{Turbine Distribution Algorithm (TDA)}
\label{alg:TurbineAlgorithm}
\end{algorithm}
\end{minipage}
\end{figure}

In order to ensure that the turbines' initial placement respects the safety distance constraint, our TDA follows the approach taken in \cite{EWEA2011}. There, the algorithm deterministically initializes the turbines in a grid formation of greatest space. The grid is constructed in such a way that the distance between the columns and rows is maximized, including the placement of turbines on the borders of the wind farm area. 
This is a straightforward approach, which is used frequently in practice, as the wake effect is already reduced to some extent (when compared to tighter layouts), even without considering the directional distribution of the wind.

Over the course of the optimization, new layouts are created based on the best-so-far  turbine configuration. For the next layout, a copy of the best-so-far is made, and a modification is then applied, in which a single turbine is selected uniformly at random and is shifted by a displacement vector $\vec v'$. In the following, we motivate our way to compute this displacement vector. 

\begin{figure}[!ht]\centering
\includegraphics[scale=0.44]{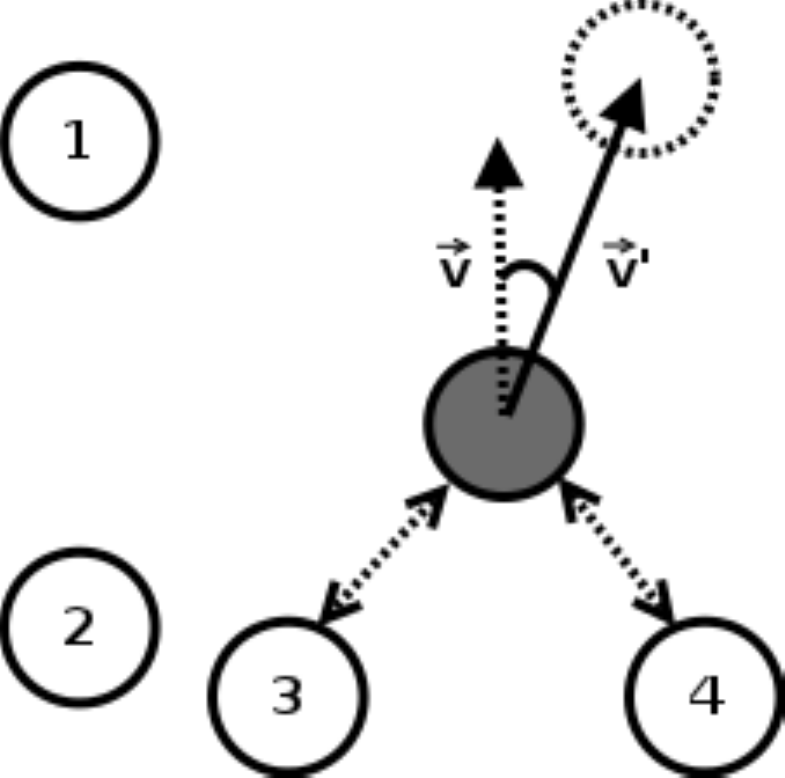}
\caption{Illustration of vector displacement using TDA with $nn=2$.}
\label{fig:NearestNeighbours}
\end{figure} 

Initially, the vector $\vec v$ is determined by the normalised sum of the difference in distance between the current turbine and its $nn$ nearest neighbours. The displacement of the current turbine by $\vec v$ would move the turbine away from the nearest neighbours. In order to ensure non-deterministic performance
within sensible limits, TDA applies a normal distribution with a fixed standard deviation to the direction of the vector and uses an adaptable standard deviation on a 0 mean over a normal distribution to determine the distance for the vector. As a superior energy output for the wind farm could be obtained by moving turbines closer together, this new vector may be randomly reversed with a preset probability.

In the next step, the new location of the turbine after displacement by the resultant vector $\vec v'$ is investigated to ensure that it would not be placed in an illegal position, currently defined as outside the wind farm bounds or too close to another turbine. If the turbine would be in an illegal position, the distance of $\vec v'$ is reduced to a legal value. An example of this displacement using $nn=2$ may be seen in Figure \ref{fig:NearestNeighbours}. Note that, in this example, the two nearest neighbours of the current turbine are turbines 3 and 4, and hence these are the only turbine locations that affect the new position of the current turbine.

At the conclusion of this layout modification, the quality of the new layout is compared against that of the best-so-far, using Equation~\ref{eq:totalenergy} and the computational speedup presented above. If the new potential solution is of higher quality, it replaces the previous best. 

Furthermore, when the quality of the layout improves due to the displacement of a turbine, the above-mentioned adaptable standard deviation for the distance of that particular turbine increases to allow for increased exploration. Similarly, when the new layout is of lower quality due to the displacement of a turbine, the standard deviation of this particular turbine decreases to allow displacement exploitation. 
Effectively, this gives TDA the ability to autonomously switch between exploration and exploitation. 
There will be local optima due to the local search properties of the algorithm, and our strategy to increase the potential of escape is the combination of $n$ adaptable displacement parameters, in combination with values drawn randomly from different normal distributions.


\section{Experimental Investigations}\label{sect:results}

In order to justify our design decisions, we performed experimental investigations on which we report in the following. 
First, we introduce the scenario that defines the wind resource present at an imaginary prospective site of a wind farm. Then, we describe TDA's parameter settings and the test scenarios for the subsequent comparative study. There, we compare the solution quality and runtime of our algorithm to a tuned version from \cite{EWEA2011} and the industry tool AWS OpenWind~\cite{openwind}.

\subsection{Algorithm Settings and Scenario Information} 

To evaluate our algorithm's performance we set up a realistic scenario with the wind resources defined as \emph{Scenario 2} in \cite{KusiakLayout2010}. The prevailing wind direction covers a broad sector of about 105$^ \circ $, and the wind intensity per direction is given by Weibull distributions. This results in non-zero probabilities for wind coming from any direction, and 
therefore, one has to optimize the layout to work with minimum wake loss along all the wind directions. 

For the subsequent experiments, the internal parameters of our algorithm were set as follows. 

The direction standard deviation is set to ${\sigma_{\{dir\}}}_i=\pi/6$ for all $i$ turbines. Thus, the resulting displacement direction will be within $\pm 30^0$ with a probability of 68.2\% of all directional adjustments, and within $\pm 60^0$ with a probability of 95.4\%. Furthermore, the reversal probability of the displacement was set to $p=0.2$. 

The displacement distance standard deviation ${\sigma_{\{dis\}}}_i$ is set and adjusted by the algorithm for each turbine individually, depending on the initial layout. If the minimal initial distance between turbines is $d$, then ${\sigma_{\{dis\}}}_i=(d-8R)/3$. Thus, we restrict the displacements to small distances, while considering the safety distance of $8R$ between the turbines. 

As described in Section~\ref{sec:tda}, these parameters are adjusted over the course of the optimization depending on whether recent displacements were successful or not. This allows for an efficient exploration as well as exploitation. 

\subsection{Impact of the Nearest Turbines}
\begin{figure}[t]
\centering
\includegraphics[width=95mm]{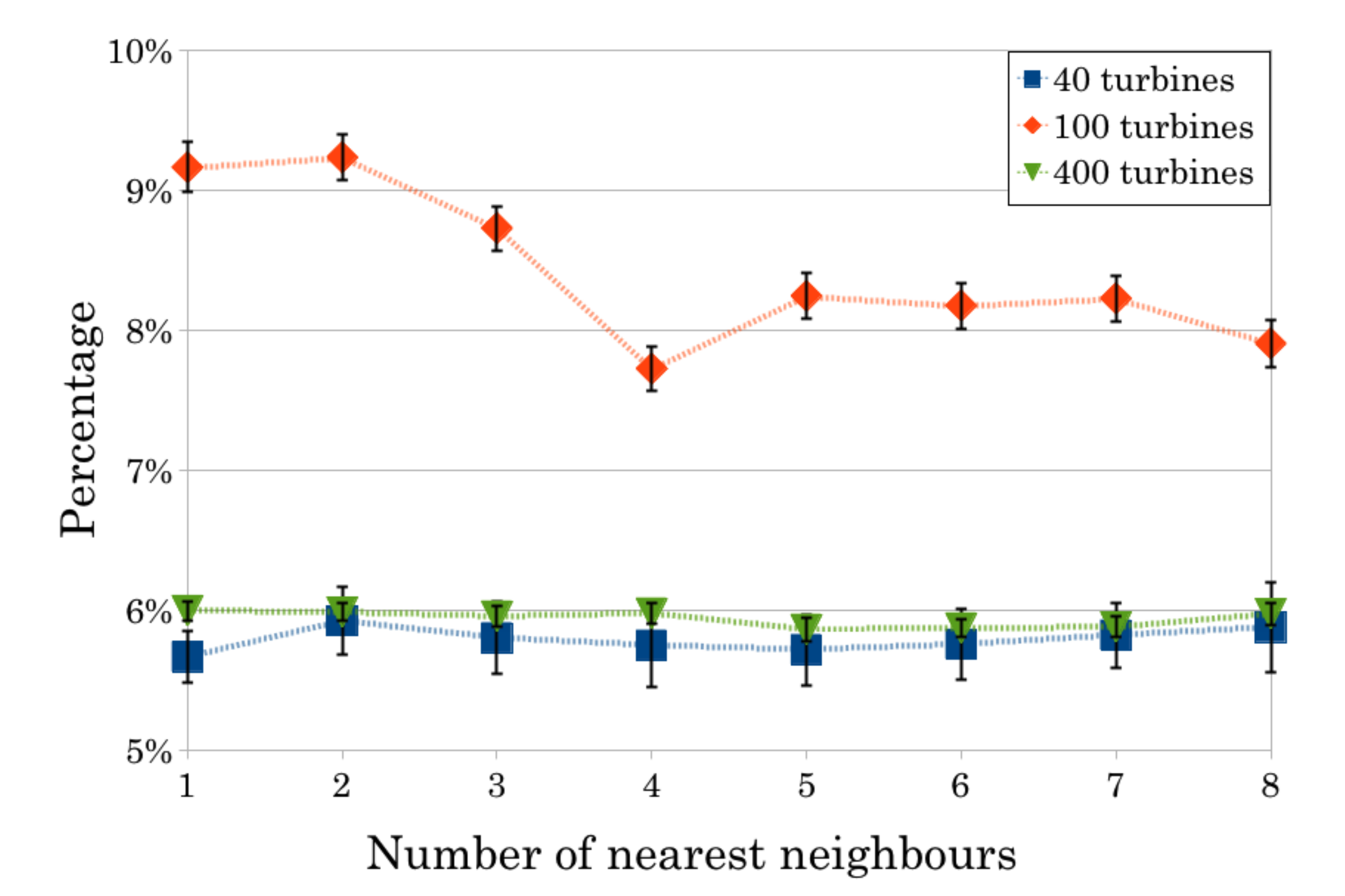}
\caption{Study: influence of the number of considered nearest neighbours on the performance. Shown is the energy gain and its standard deviation achieved over the initial layout.
}
\label{fig:nnstudy}
\end{figure}

In order to understand the influence of the number of nearest neighbouring turbines $nn$ that influence the relocation of a chosen turbine, we ran an initial set of experiments. We chose the scenarios with 40, 100, and 400 turbines, and ran each experiment with a budget of 10,000 evaluations. $nn$ varied from 1 to 8, and the results of 100 repetitions per scenario are shown in Figure~\ref{fig:nnstudy}.

We noticed that the number of nearest neighbours sometimes has some influence on the performance of the algorithm. Depending on the scenario, a smaller value of $nn$ seems to be beneficial. For example for the 100 turbines scenarios, using $nn=2$ instead of $nn=8$ yielded in layouts with an increased energy output of 1.1\% on average. For the 40 turbine scenario, however, the maximum distance in the averages is only 0.6\%, and for the 400 turbines just 0.1\%. A possible explanation is that with four and eight neighbours, the influence of two opposing neighbouring turbines is cancelled out: the resulting displacement vector for a selected turbine just has a `unspecific' direction, instead of an `explorative' direction as is the case when only one or two neighbours are considered.

\subsection{Experimental Results}

To assess the performance of our algorithm, we compare it directly with the results from \cite{EWEA2011} and the approach implemented in AWS OpenWind. In our computational study, we set up several scenarios with varying numbers of turbines, and varying farm sizes. For $n=10, 20, \ldots, 100$ a quadratic farm of $3\times3km^2$ was chosen, and for $n=200, 300, 400, 500, 1000$ rectangular farms of $8\times5km^2$, $10\times6km^2$, $12\times6km^2$, $14\times7km^2$, and $20\times10km^2$. 

To allow for a better comparison with the CMA-ES algorithm presented in \cite{EWEA2011}, where 10,000 generations with a total of 200,000 evaluations were performed, we ran our algorithm for 10,000 and 200,000 evaluations---we stopped the algorithms if no improvement was reached after 1,000 evaluations. In order to compare our TDA not only to an academic system, we additionally implemented the optimizer component that is part of the industrial tool AWS OpenWind, and ran it for a maximum of 200,000 evaluations. 
Quickly, we noticed that given the wide range of scenarios, the original CMA-ES fails to produce consistent results---in particular, it performed poorly for 70 and 80 turbines. Despite the fact that it has elaborate adjustment capabilities, it is not always capable of learning the problem-specific features of the solution space within reasonable run-times. Through the increase of CMA-ES's initial standard deviation, the explorative phase was extended, and we were able to increase its quality performance on these two particular scenarios to a level comparable with TDA. In the following, we refer to this tuned version as CMA-ES*. 

{ \renewcommand{\tabcolsep}{0.59mm}
\renewcommand{\arraystretch}{1.25}
\begin{table}
\caption{Results. Reported are the predicted energy outputs in kW (average and maximum values, and the standard deviation), time in hours. The algorithm with the best average energy output has been highlighted in bold, the second-best in italics.}
\small{
\centering
\begin{tabular}{|cc|ccc|ccc|} \hline \hline
 &  & 
\multicolumn{3}{c|}{CMA-ES* 200k} & 
\multicolumn{3}{c|}{OpenWind 200k} \\
n & Initial
 & avg$_\mathrm{stdev}$ & max & time & avg$_\mathrm{stdev}$ & max & time  
\\ \hline
10 & 7.290E+4 & \textit{7.306E+4}$_{3.63E+1}$  & 7.315E+4 & 0.10 & 7.290E+4$_{0.00E+0}$ & 7.290E+4 & 0.01 \\
20 & 1.448E+5 & \textbf{1.448E+5}$_{0.00E+0}$ & 1.448E+5 & 0.05 & \textbf{1.448E+5}$_{0.00E+0}$ & 1.448E+5 & 0.01  \\
30 & 2.015E+5 & 2.096E+5$_{1.73E+2}$ & 2.100E+5 & 1.29 & 2.052E+5$_{1.17E+3}$ & 2.067E+5 & 0.01  \\
40 & 2.630E+5 & 2.742E+5$_{2.07E+2}$ & 2.746E+5 & 2.01 & 2.671E+5$_{8.08E+2}$ & 2.688E+5 & 0.02  \\
50 & 3.247E+5 & 3.367E+5$_{3.01E+2}$ & 3.372E+5 & 3.82 & 3.269E+5$_{7.28E+2}$ & 3.282E+5 & 0.03  \\
60 & 3.688E+5 & 3.947E+5$_{3.47E+2}$ & 3.953E+5 & 5.45 & 3.837E+5$_{1.19E+3}$ & 3.858E+5 & 0.04  \\
70 & 4.341E+5 & 4.435E+5$_{8.98E+3}$ & 4.559E+5 & 5.90 & 4.370E+5$_{9.03E+2}$ & 4.389E+5 & 0.07  \\
80 & 4.707E+5 & \textit{5.077E+5}$_{1.02E+3}$ & 5.098E+5 & 9.78 & 4.884E+5$_{1.18E+3}$ & 4.906E+5 & 0.08  \\
90 & 5.207E+5 & 5.541E+5$_{1.22E+3}$ & 5.567E+5 & 9.64 & 5.333E+5$_{1.02E+3}$ & 5.356E+5 & 0.10  \\
100 & 5.535E+5 & 5.994E+5$_{2.18E+3}$ & 6.029E+5 & 14.0 & 5.762E+5$_{1.48E+3}$ & 5.784E+5 & 0.14  \\
200 & 1.248E+6 & 1.301E+6$_{4.36E+2}$ & 1.302E+6 & 40.1 & 1.273E+6$_{2.63E+3}$ & 1.278E+6 & 0.39  \\
300 & 1.861E+6 & 1.935E+6$_{7.76E+2}$ & 1.937E+6 & 111 & 1.893E+6$_{2.77E+3}$& 1.898E+6 & 0.87  \\
400 & 2.415E+6 & 2.549E+6$_{1.02E+3}$ & 2.550E+6 & 221 & 2.480E+6$_{2.20E+3}$ & 2.485E+6 & 1.46  \\
500 & 3.062E+6 & 3.196E+6$_{9.63+E3}$ & 3.201E+6 & 324 & 3.118E+6$_{3.55E+3}$ & 3.124E+6 & 2.26  \\
1000 & 6.023E+6 & 6.298E+6$_{2.36E+4}$ & 6.335E+6 & 327 & 6.202E+6$_{4.81E+3}$ & 6.210E+6 & 9.43  \\
\hline 
\end{tabular}

\renewcommand{\tabcolsep}{0.25mm}
\begin{tabular}{|c|ccc|ccc|cc|} \hline 
 & 
\multicolumn{3}{c|}{TDA 10k} & 
\multicolumn{5}{c|}{TDA 200k} \\
n & avg$_\mathrm{stdev}$  & max & time & avg$_\mathrm{stdev}$  & max & time & 
$P_{\mathit{loss}}$ &
$P_{\mathit{gain}}$ 
\\ \hline
10 & 7.305E+4$_{1.73E+1}$ & 7.308E+4 & 0.01 & \textbf{7.309E+4}$_{2.47E+1}$ & 7.314E+4 & 0.22 & 0.1\% & 100.0\%\\
20  & \textbf{1.448E+5}$_{7.05E-1}$ & 1.448E+5 & 0.02 & \textbf{1.448E+5}$_{1.30E+1}$ & 1.449E+5 & 0.50 & 1.0\% & 98.1\%\\
30  & \textit{2.123E+5}$_{3.74E+2}$ & 2.131E+5 & 0.04 & \textbf{2.135E+5}$_{4.08E+2}$ & 2.144E+5 & 0.75 & 2.7\% & 92.2\%\\
40 & \textit{2.772E+5}$_{4.46E+2}$ & 2.781E+5 & 0.05 & \textbf{2.791E+5}$_{5.75E+2}$ & 2.806E+5 & 1.02 & 4.6\% & 88.7\%\\
50 & \textit{3.392E+5}$_{4.61E+2}$ & 3.401E+5 & 0.06 & \textbf{3.412E+5}$_{3.42E+2}$ & 3.418E+5 & 1.29 & 6.7\% & 84.8\%\\
60  & \textit{3.980E+5}$_{5.25E+2}$ & 3.991E+5 & 0.08 & \textbf{4.011E+5}$_{5.25E+2}$ & 4.022E+5 & 1.58 & 8.6\% & 80.4\%\\
70  & \textit{4.512E+5}$_{1.19E+3}$ & 4.537E+5 & 0.09 & \textbf{4.555E+5}$_{1.57E+3}$ & 4.591E+5 & 1.86 & 11.1\% & 72.8\%\\
80 & 5.044E+5$_{1.15E+3}$ & 5.083E+5 & 0.11 & \textbf{5.090E+5}$_{1.01E+3}$ & 5.108E+5 & 2.15 & 13.0\% & 72.8\%\\
90 & \textit{5.548E+5}$_{1.34E+3}$ & 5.569E+5 & 0.12 & \textbf{5.609E+5}$_{9.38E+2}$ & 5.625E+5 & 2.52 & 14.8\% & 68.9\%\\
100 & \textit{6.015E+5}$_{1.32E+3}$ & 6.041E+5 & 0.14 & \textbf{6.083E+5}$_{1.28E+3}$ & 6.113E+5 & 2.83 & 16.9\% & 63.9\%\\
200  & \textit{1.309E+6}$_{8.99E+2}$ & 1.311E+6 & 0.33 & \textbf{1.323E+6}$_{9.16E+2}$ & 1.325E+6 & 6.73 & 9.6\% & 96.8\%\\
300 & \textit{1.949E+6}$_{1.37E+3}$ & 1.952E+6 & 0.55 & \textbf{1.971E+6}$_{9.72E+2}$ & 1.973E+6 & 11.3 & 10.2\% & 87.4\%\\
400  & \textit{2.553E+6}$_{1.52E+3}$ & 2.557E+6 & 0.84 & \textbf{2.584E+6}$_{1.12E+3}$ & 2.586E+6 & 17.0 & 11.7\% & 82.6\%\\
500 &  \textit{3.211E+6}$_{1.66E+3}$ & 3.215E+6 & 1.19 & \textbf{3.249E+6}$_{1.55E+3}$ & 3.251E+6 & 24.5 & 11.2\% & 90.1\%\\
1000 & \textit{6.363E+6}$_{2.49E+3}$ & 6.368E+6 & 3.69 & \textbf{6.449E}+6$_{2.04E+3}$ & 6.454E+6 & 75.0 & 11.9\% & 86.2\%\\
\hline 
\hline 
\end{tabular}
}
\label{tab:results}
\end{table}%
}

\begin{figure}
\centering
\includegraphics[width=100mm]{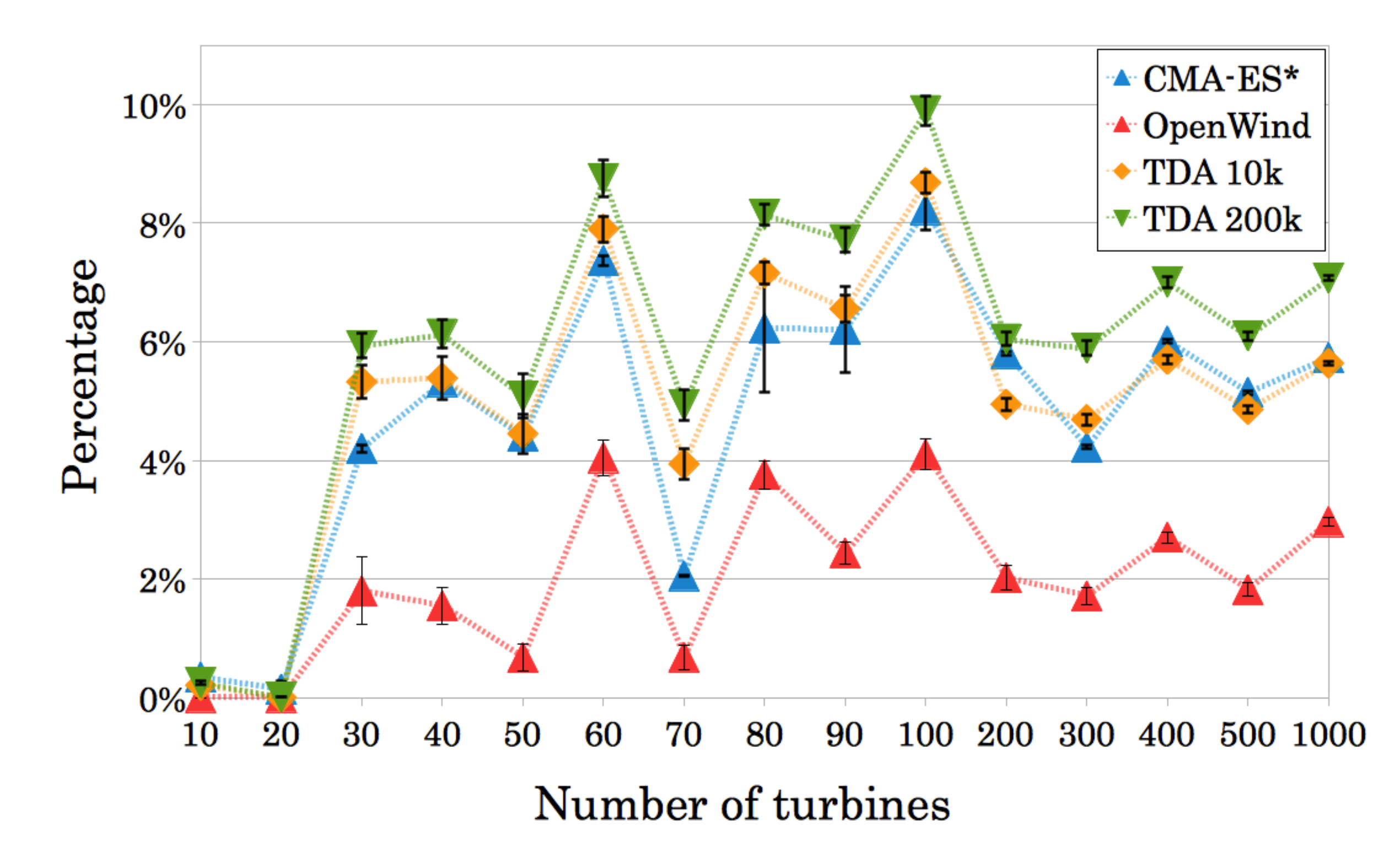}
\caption{Performance comparison: our algorithm TDA (using 10,000 and 200,000 evaluations) versus
the tuned academic CMA-ES* and the industry tool OpenWind (both using 200,000 evaluations). Shown is the energy gain achieved over the initial layout, and the standard deviation over the 30 repetitions.}
\label{fig:comparison}
\end{figure}

The initial layout, where turbines are placed in a maximally spaced grid, was identical for all algorithms. 
Each scenario was repeated 30 times, and the results are listed in Table~\ref{tab:results} and in Figure~\ref{fig:comparison}.\footnote{In the scenarios with 10 and 20 turbines, the turbines are very far away from each other, with just minimal wake effects occurring.} 
As can be seen, our algorithm's results (when using only 10,000 evaluations) are comparable to those of CMA-ES*, and in many cases have a higher solution quality. Additionally, our algorithm just uses a fraction of the evaluations and of the time to reach the results. 
Exemplarily, the academic CMA-ES* needs $2h$ for the 40 turbines scenario, while even a better gain is reached within just three minutes by our algorithm. And for the 400 turbines, our algorithm needs less than an hour to achieve the results for which CMA-ES* requires $>200h$. 
If given 20 times the number of evaluations, our algorithm is able to produce layouts that produce up to 1.4\% more energy than the CMA-ES* based algorithm from \cite{EWEA2011}. In addition, it is significantly faster. 

Independent of the scenario, OpenWind's optimizer achieves an average improvement over the initial layouts of only 2\%. Responsible for this is most likely the lack of self-adaptation, as the optimizer gets stuck in local optima. This happens very quickly: typically within the first 1,000 layout assessments, which is reflected in the short running times. Contrary to this, our TDA adapts the relocation parameters for each turbine independently of the others. The effects of this ability to change from exploration to exploitation and back is reflected in the large performance advantage of TDA over OpenWind.

To the results, we added some derived statistics. $P_{\mathit{loss}} = \frac{\mathit{wake\;loss}}{\mathit{energy\;captured}}$ denotes the average percentage of unused wind energy due to wake effects. As expected, this loss increases for the scenarios with 10-100 turbines, as more turbines are packed into the same area, inflicting increased wake losses. For 200-1000 turbines, this value is relatively constant, as the chosen sizes of the farms vary such that the farms could contain roughly $2n$ turbines. 
$P_{\mathit{gain}}$ is the average percentage in energy gained over the previous scenario (i.e. after increasing the number of turbines by 10, 100, or 500). Again, the effect of the increased mutual wakes is reflected in a decreasing benefit of adding turbines, when compared to the next smaller scenario.


\section{A Real-World Problem: Dealing With Infeasible Areas}\label{sec:infeasibleAreas}

\begin{figure}%
\centering
\includegraphics[height=52mm]{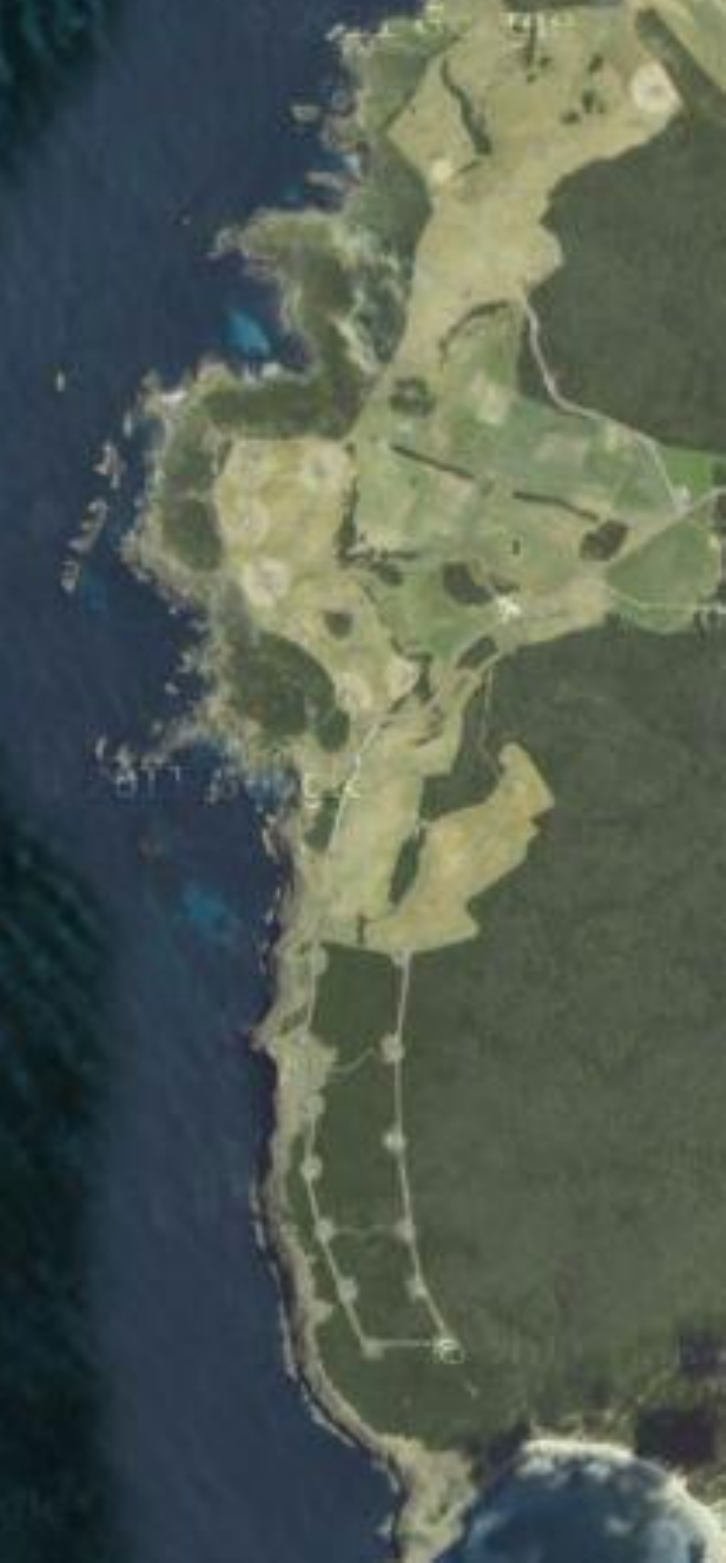}
\includegraphics[height=52mm]{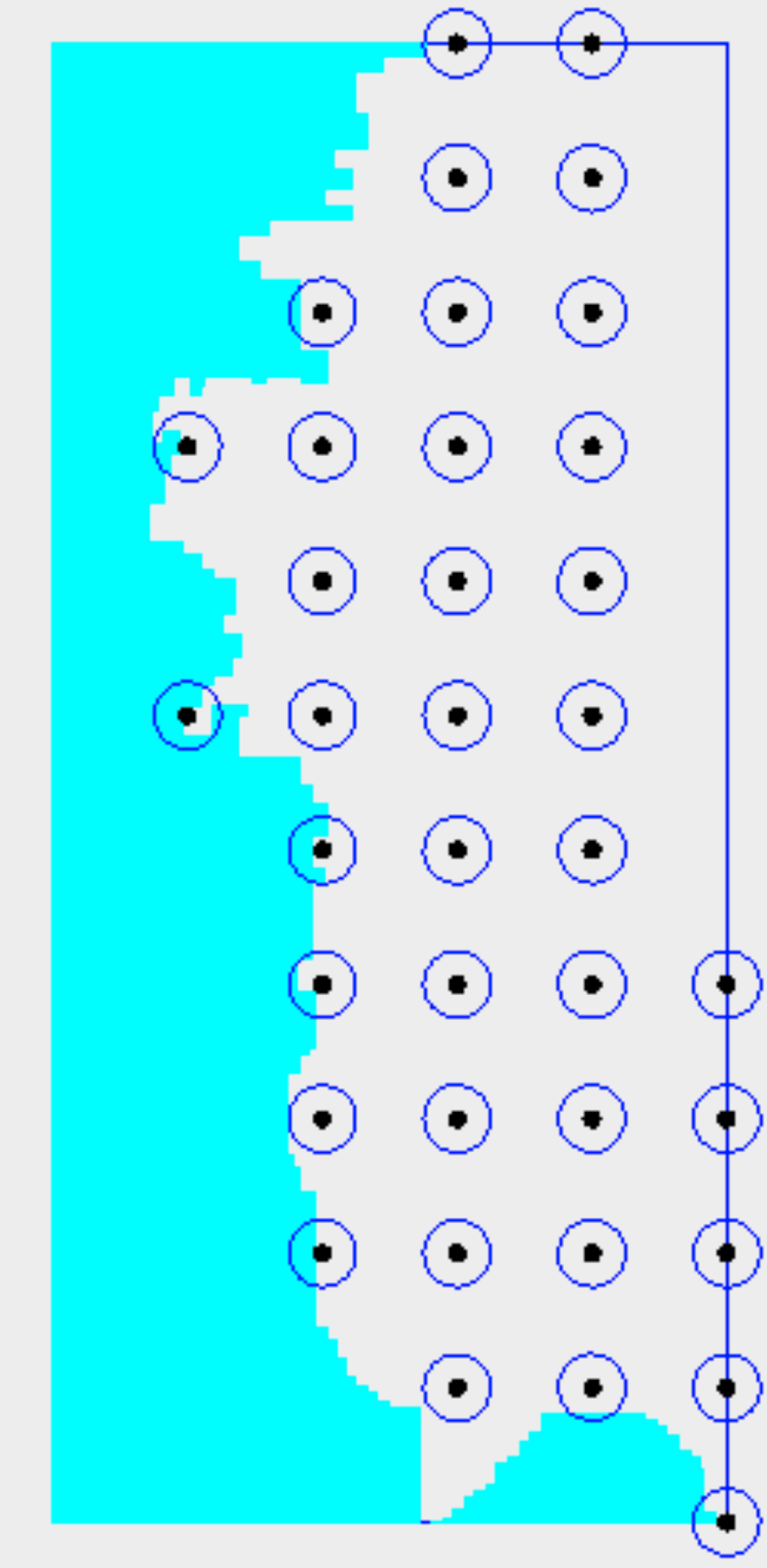}
\includegraphics[height=52mm]{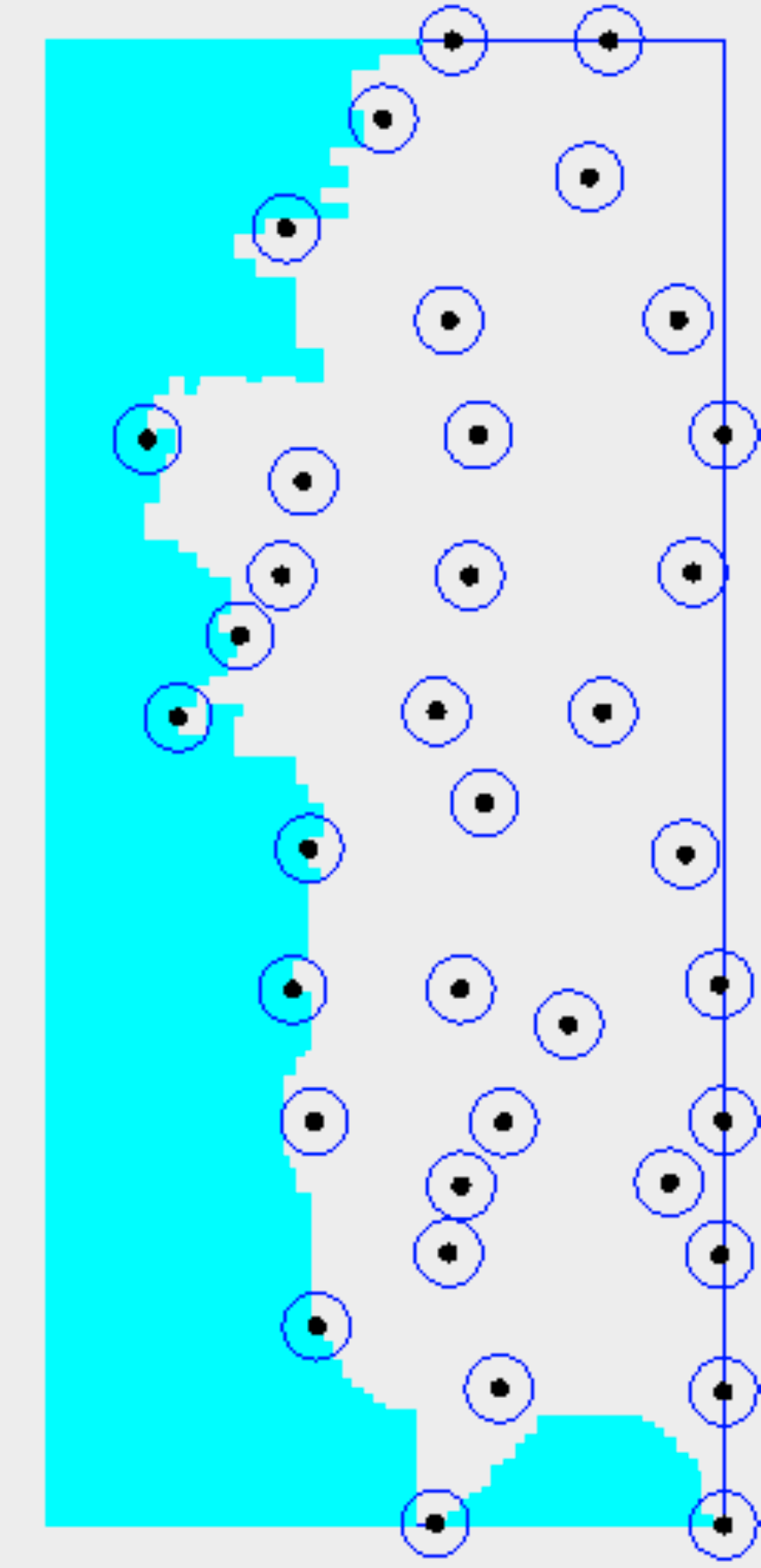}
\includegraphics[height=52mm]{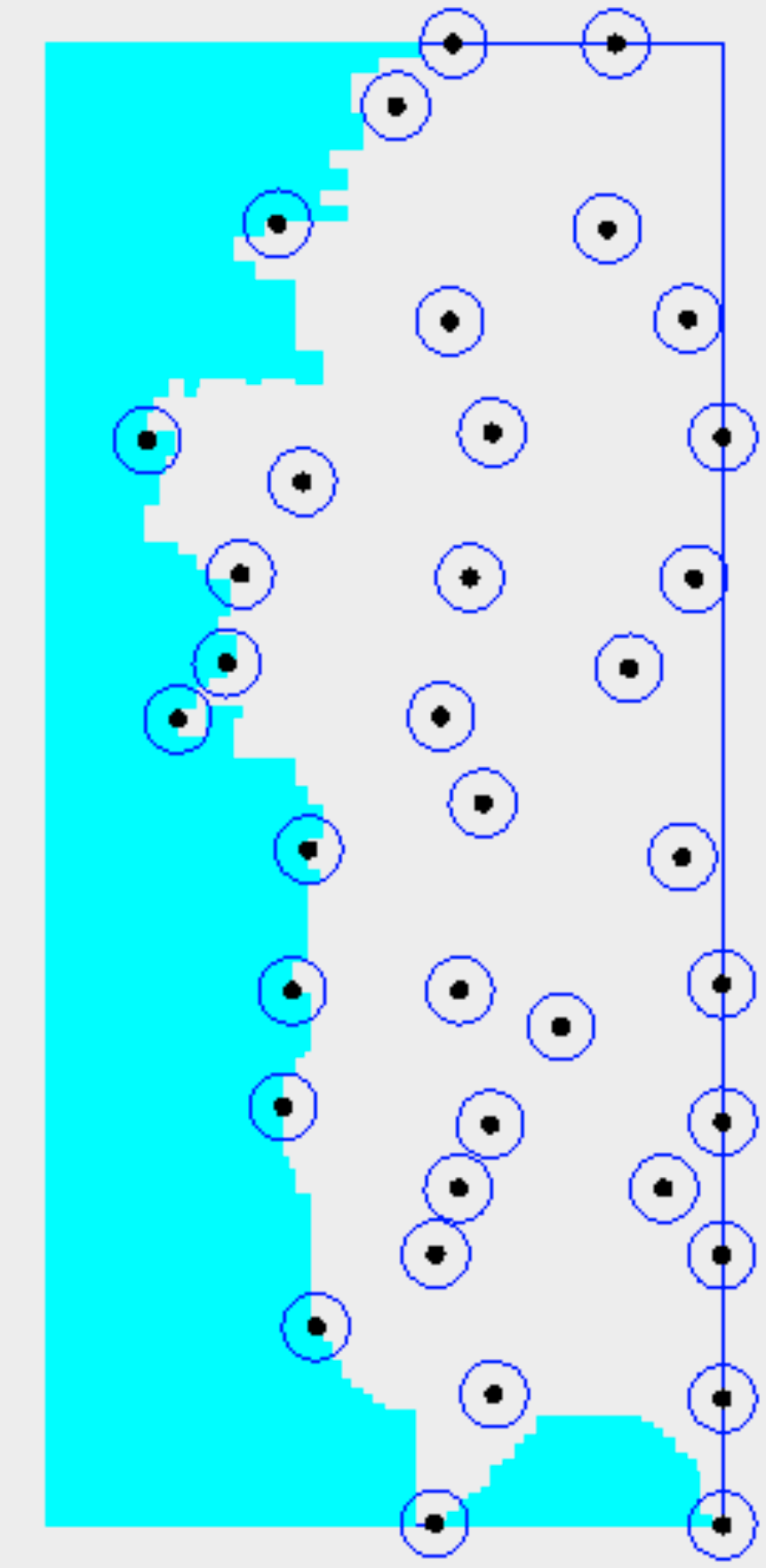}
\caption{Demonstration of the TDA infeasible area modelling capability. The image on the left portrays a satellite image of the Woolnorth wind farm in Tasmania, Australia~\protect\cite{google}. The right images are examples of the the loose adaptation used in the modelling tool, and from left to right model the scenario at 0 (252.9 MW), 5,000 (264.91 MW), and 20,000 (265.8 MW) evaluations.}%
\label{fig:realWorldScenario}%
\end{figure}

It must be noted that the layout of wind farms are generally unable to be denoted simply by a rectangular area specified only by a width and height. Actual wind farms may have uneven boundaries due to proximity to unstable ground, lakes, cliffs, or may simply not have authority or ownership to build in certain locations. These and other geographical constraints may additionally exist within the bounds of the wind farm. Any algorithm that attempts to realistically model a wind farm must model these constraints.

TDA has been adapted to model these infeasible areas, adding an extra constraint to those specified in Section \ref{sec:constraints}. By modelling these infeasible regions as sets of shapes, any operation which would move a turbine into an infeasible region is corrected in the same manner in which the proximity and farm area constraints are handled above: 
when the application of the displacement vector $\vec v'$ (line 13 of Algorithm \ref{alg:TurbineAlgorithm}) would place the turbine in an infeasible zone, the distance of $\vec v'$ is reduced until the displacement is legal.

This extension has resulted in a capable wind farm modelling tool. In order to demonstrate the ability of the algorithm to model real-world wind farms, Figure~\ref{fig:realWorldScenario} presents both the satellite image and a loosely modelled representation of the Woolnorth wind farm in Tasmania, Australia (40.685$^{\circ}$S 144.717$^{\circ}$E). 
The modelled scenario uses the 37 turbines contained in the north Woolnorth site. However this scenario is merely a proof of concept, as it is not using the wind characteristics of Woolnorth, nor specific internal map or terrain information of the actual site. 
Only the coastal details have been represented, and the above-mentioned wind resource is used (\emph{Scenario 2} in \cite{KusiakLayout2010}).

As can be seen, the turbines observe the constraints of the problem and are quickly distributed into a superior formation from evaluation 0 through to evaluation 20,000. As the wind is predominantly from the western direction (between 120$^ \circ $ to 225$^ \circ $), the turbines tend to form in staggered north/south columns while leaving space along the east/west directions.


\section{Conclusion}\label{sect:conclusion}

In this paper, we have presented a fast and efficient algorithm for the layout optimization of large wind farms. It takes problem-specific features into account, and benefits from the achieved reduced computational complexity of a layout evaluation when considering the Park wake model. As a result, our algorithm achieves higher quality results than existing approaches, while the assessment speed-up allows for an optimization within minutes or hours instead of days or weeks (effective speed-up factors of up to 270 were observed). 
As the parallelization of the neighbourhood investigations is natural with most local search approaches, 
we expect that a parallelization of our approach can yield further speedups.

Although we considered one specific wake model, namely the Park wake model, it is important to note that our optimization algorithm can be easily applied to other wake models such as the deep array wake model~\cite{dawmBarth}.

\end{document}